\documentclass[sigconf]{acmart}
\AtBeginDocument{%
  }

\setcopyright{acmlicensed}
\copyrightyear{2018}
\acmYear{2018}
\acmDOI{XXXXXXX.XXXXXXX}
\acmConference[Conference acronym 'XX]{Make sure to enter the correct
  conference title from your rights confirmation email}{June 03--05,
  2018}{Woodstock, NY}
\acmISBN{978-1-4503-XXXX-X/2018/06}

\copyrightyear{2025}
\acmYear{2025}
\setcopyright{acmlicensed}\acmConference[MMSports '25]{Proceedings of the 8th International ACM Workshop on Multimedia Content Analysis in Sports}{October 27--28, 2025}{Dublin, Ireland}
\acmBooktitle{Proceedings of the 8th International ACM Workshop on Multimedia Content Analysis in Sports (MMSports '25), October 27--28, 2025, Dublin, Ireland}
\acmDOI{10.1145/3728423.3759408}
\acmISBN{979-8-4007-1835-9/2025/10}

\usepackage{subcaption}
\usepackage{booktabs}
\usepackage{multirow}
\usepackage{booktabs}  % 导言区添加
\usepackage{pifont}
\usepackage{balance}

\begin{document}

\title{Shot2Tactic-Caption: Multi-Scale Captioning of \\ Badminton Videos for Tactical Understanding}

\author{Ning Ding}
\affiliation{%
  \institution{Nagoya Institute of Technology}
  \city{Nagoya}
  \country{Japan}}
\email{ding.ning@nitech.ac.jp}

\author{Keisuke Fujii}
\affiliation{%
  \institution{Nagoya University}
  \city{Nagoya}
  \country{Japan}}
\email{fujii@i.nagoya-u.ac.jp}

\author{Toru Tamaki}
\affiliation{%
 \institution{Nagoya Institute of Technology}
 \city{Nagoya}
 \country{Japan}}
\email{tamaki.toru@nitech.ac.jp}


\begin{abstract}
Tactical understanding in badminton involves interpreting not only individual actions but also how tactics are dynamically executed over time.
In this paper, we propose \textbf{Shot2Tactic-Caption}, a novel framework for semantic and temporal multi-scale video captioning in badminton, capable of generating shot-level captions that describe individual actions and tactic-level captions that capture how these actions unfold over time within a tactical execution.
We also introduce the Shot2Tactic-Caption Dataset, the first badminton captioning dataset containing 5,494 shot captions and 544 tactic captions. 
Shot2Tactic-Caption adopts a dual-branch design, with both branches including a visual encoder, a spatio-temporal Transformer encoder, and a Transformer-based decoder to generate shot and tactic captions.
To support tactic captioning, we additionally introduce a Tactic Unit Detector that identifies valid tactic units, tactic types, and tactic states (e.g., Interrupt, Resume).
For tactic captioning, we further incorporate a shot-wise prompt-guided mechanism, where the predicted tactic type and state are embedded as prompts and injected into the decoder via cross-attention. 
The shot-wise prompt-guided mechanism enables our system not only to describe successfully executed tactics but also to capture tactical executions that are temporarily interrupted and later resumed.
Experimental results demonstrate the effectiveness of our framework in generating both shot and tactic captions. Ablation studies show that the ResNet50-based spatio-temporal encoder outperforms other variants, and that shot-wise prompt structuring leads to more coherent and accurate tactic captioning.
\end{abstract}

\ccsdesc[500]{Computing methodologies~Video summarization}
\ccsdesc[500]{Computing methodologies~Activity recognition and understanding}
\ccsdesc[500]{Computing methodologies~Natural language generation}
\ccsdesc[300]{Applied computing~Computers in other domains}

\keywords{Video Captioning, Sports Video Analysis, Datasets, Racket sports}



\maketitle

\section{Introduction}

In competitive sports, the game situation constantly evolves, driven by rapid and complex player actions and team interactions. Understanding such dynamic contexts requires not only perceiving the visual cues but also interpreting the underlying tactical intentions. 
To support such understanding, video captioning has gained attention in sports as a promising technique for describing match situations in natural language, thereby facilitating human comprehension of tactics and context.

Among various sports, racket sports are characterized by extremely short events (i.e., shots) and continuous event sequences. Rather than occurring in isolation, events typically occur consecutively and form a sequence that constitutes a coherent tactical unit.
However, most existing studies have focused on soccer, where Dense Video Captioning (DVC) \cite{krishna2017dense, iashin2020multi, zhou2024streaming} is used to detect and describe discrete events such as goals and passes. While previous studies \cite{yu2018fine, HAMMOUDEH2022104, qi2023goal, mkhallati2023soccernet, suglia2022going} can generate captions for isolated events, they struggle to capture how sequential events contribute to higher-level tactical structures.

To address these challenges, we propose a multi-scale captioning framework that simultaneously generates concise shot captions as well as high-level tactic captions that explain how actions unfold within a tactic. Our goal is to move beyond flat sequences of isolated actions and toward a context-aware interpretation of tactical structure and execution.

Furthermore, in real matches, tactics rarely unfold exactly as planned. Especially in top-level matches, players are constantly engaged in tactical confrontations, attempting to execute tactics while opponents try to disrupt or counter them. Therefore, tactical execution is a dynamic process, where a tactic may be interrupted mid-way and later resumed during play.
Therefore, to capture tactical execution in a realistic and robust manner, it is essential to explicitly model the dynamic flow of tactics: how they begin, progress, get interrupted, resume, and finish. In this study, we define the tactical process using 5 states: ``Start'', ``Continue'', ``Interrupt'', ``Resume'', and ``Finish''. 
Specifically, ``Start'' marks the initiation of a tactic, ``Continue'' indicates that the tactic is actively unfolding, ``Interrupt'' denotes an unexpected disruption, ``Resume'' captures the reactivation of a previously interrupted tactic, and ``Finish'' signals its completion.
This allows our system to not only describe successfully executed tactics but also to capture tactical executions that are interrupted and later resumed, thereby enabling a more comprehensive and realistic understanding of tactical execution.

Besides, although several datasets with event-level captions have been released for sports like soccer and basketball \cite{yu2018fine, mkhallati2023soccernet, HAMMOUDEH2022104}, they do not provide aligned annotations that capture both fine-grained actions and their organization into coherent tactical processes.
To fill this gap, We built the Shot2Tactic-Caption Dataset, the first badminton doubles captioning dataset containing aligned shot-level and tactic-level annotations.

The key contributions of this study are summarized as follows:
    \begin{itemize}
    \item We introduce the Shot2Tactic-Caption dataset, the first multi-scale captioning dataset for badminton, providing both shot captions and tactic captions.
    
    \item We propose Shot2Tactic-Caption, a novel framework for multi-scale video captioning in badminton, capable of generating both shot captions and tactic captions that describe how actions unfold within a tactic.
    
    \item We introduce a prompt-guided captioning mechanism that uses predicted tactic types and predefined tactic states (e.g., Interrupt, Resume) as inputs to model the temporal progression and dynamic transitions of tactical execution.
\end{itemize}

% \begin{table*}[t]
% \centering
% % \footnotesize  % 使用默认字号
% \begin{tabular}{l|p{3.5cm}|r|r|r|c}
% \hline
% \textbf{Name} & \textbf{Coverage} & \textbf{Video} & \textbf{Total Duration (min)} & \textbf{Captions} & \textbf{Cap/min} \\
% \hline

% Soccer Captioning~\cite{HAMMOUDEH2022104} & Key Events & 942 & 2735.9 & 22,000 & – \\

% GOAL (Suglia \textit{et al.}~\cite{suglia2022going}) & Highlights & 1,100 & 238 & 53,000 & – \\

% GOAL (Qi \textit{et al.}~\cite{qi2023goal}) & Key Events & 1,100 & 73.1 & 22,000 & 0.28 \\

% SoccerNet-Caption~\cite{mkhallati2023soccernet} & Key Events & 942 & 2735.9 & 36,894 & 0.87 \\
% \hline
% Shot2Tactic-Caption (Ours) & All Events (Every Shot) & 10 & 75.61 & 5,494 & 73 \\
% \hline
% \end{tabular}
% \caption{Comparison between Shot2Tactic-Caption dataset and other sports captioning datasets.}
% \label{tab:dataset_comparison}
% \end{table*}
%

\section{Related Work} \label{related-works}
\subsection{Sports Video captioning}

Early studies on sports video captioning have primarily focused on describing observable actions and events. 

For basketball, Fine-grained Sports Narrative~\cite{yu2018fine} focused on generating detailed narratives describing player movements and interactions. Although it captures fine-grained actions, it does not explicitly consider higher-level tactics.

In soccer, Soccer Captioning~\cite{HAMMOUDEH2022104} constructed a dataset of paired video clips and captions and proposed a multimodal model integrating image frames, optical flow, and inpainting features. However, such methods mainly produce coarse descriptions of game events and often rely on clip-caption alignment without capturing the underlying tactical dynamics.
More recently, GOAL~\cite{qi2023goal} introduced Knowledge-grounded Video Captioning (KGVC), combining video clips with curated knowledge triples to generate semantically richer and more informative commentary. While GOAL demonstrates the potential of incorporating structured knowledge to enhance captioning, it primarily targets soccer and emphasizes factual background information rather than fine-grained tactical interpretation.
Meanwhile, SoccerNet-Caption~\cite{mkhallati2023soccernet} utilized automatic speech recognition (ASR) to transcribe real-time match commentary and synchronize it with video events. This provides a valuable benchmark for learning the style of live sports broadcasting. Nonetheless, these transcriptions still tend to focus on describing isolated events and lack detailed modeling of evolving tactics or team coordination.

While significant progress has been made in generating captions from sports videos, most existing methods still focus primarily on describing individual events. They rarely attempt to explain how these sequential events form tactical executions. 
This limitation is particularly pronounced in racket sports such as badminton. Without such interpretative descriptions, badminton can appear to be merely a series of back-and-forth shots, when in reality, these sequences constitute carefully organized tactics. In our study, we propose a multi-scale captioning approach that generates captions for both individual shots and higher-level tactic units simultaneously. This joint modeling enables the clear capture of tactical flow and the generation of richer, tactic-aware descriptions.

\subsection{Datasets for Sports Video Understanding}

Most publicly available datasets for video captioning in sports have focused on team sports like soccer and basketball. For instance, the Fine-grained Sports Narrative dataset provides 6,520 captions paired with basketball gameplay clips from NBA 2K~\cite{yu2018fine}. In soccer, Suglia et al. \cite{suglia2022going} constructed a dataset consisting of 1,100 highlight videos and over 53K sentences. The Soccer Captioning dataset~\cite{HAMMOUDEH2022104} includes 22k caption-clip pairs, while the GOAL dataset~\cite{qi2023goal} provides 22k sentences describing soccer action clips with corresponding natural language captions. The recent SoccerNet-Caption dataset~\cite{mkhallati2023soccernet} contains 36,894 captions synchronized with 715.9 hours of soccer match footage.

For other sports, although SportsQA~\cite{li2024sports} provides a dataset containing videos and multiple-choice questions covering sports such as tennis, baseball, and swimming, it focuses mainly on discrete question answering and does not support continuous event description in natural language.
Among other sports, in particular, racket sports have very limited publicly available datasets for video understanding. Although a few works have released datasets for shuttle tracking \cite{9302757, pingali2000ball}, shot type recognition\cite{yenduri2024adaptive}, pose estimation \cite{WANG2024110665} and player evaluation\cite{ding2024estimation, 9775086,seong2024multisensebadminton,ban2022badmintondb}, there are currently no publicly available datasets that support caption generation for racket sports.

\begin{figure}[t]
    \centering
    \begin{subfigure}[h]{0.85\linewidth}
        \centering
        \includegraphics[width=\linewidth]{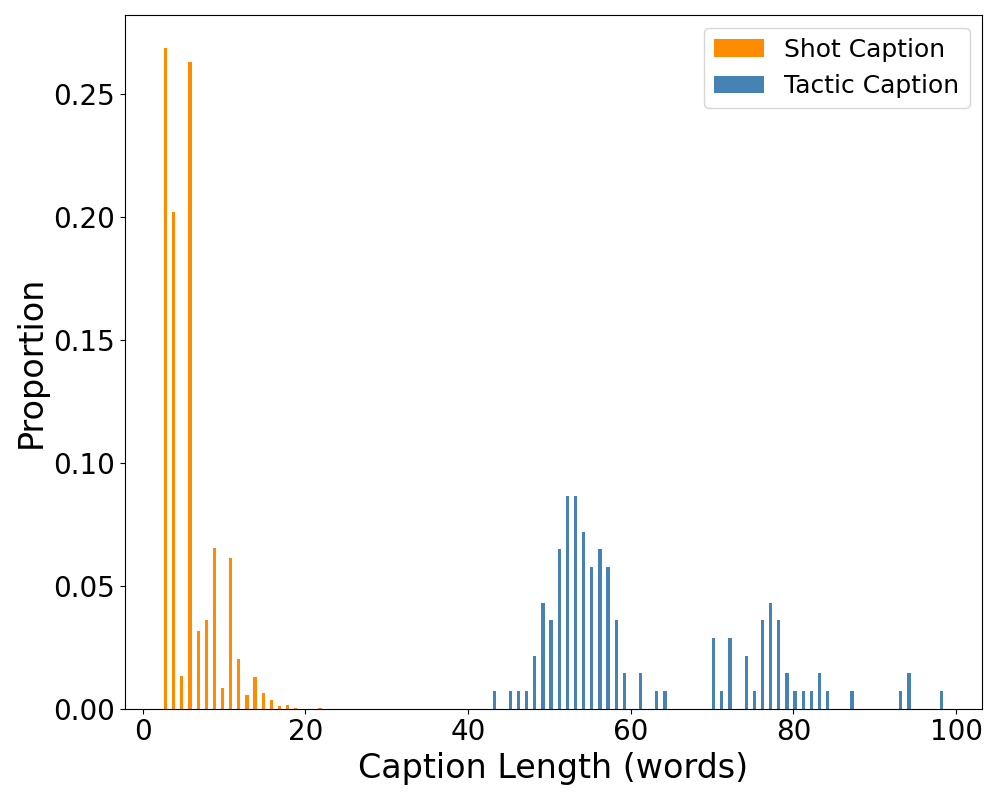}
        \caption{}
        \label{fig:caption_length_distribution}
    \end{subfigure}
    
    \vspace{1em}
    
    \begin{subfigure}[h]{0.90\linewidth}
        \centering
        \includegraphics[width=1\linewidth]{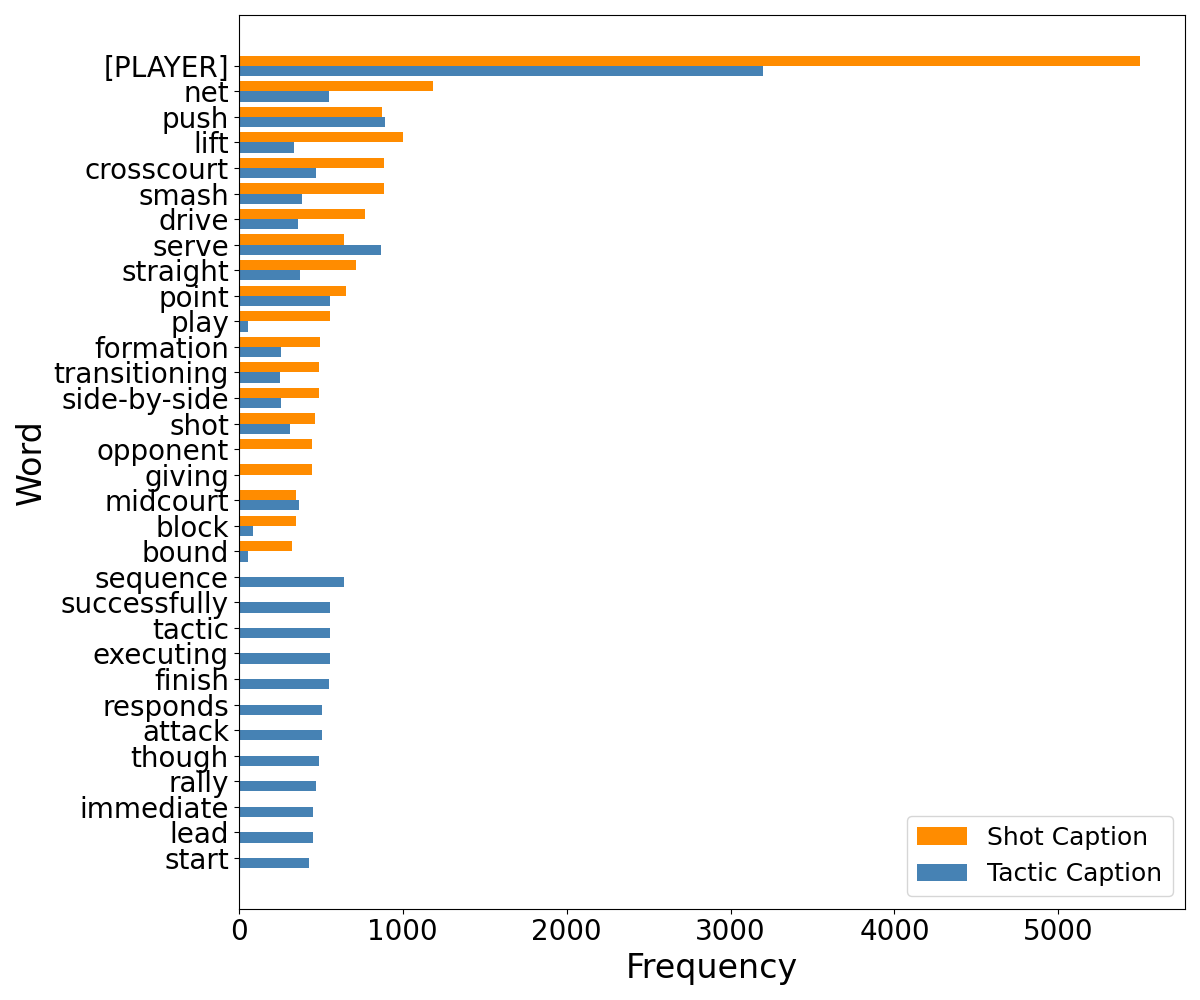}
        \caption{}
        \label{fig:most_common_words_cleaned}
    \end{subfigure}
    
    \caption{Statistical analysis of captions:
    (a) Distribution of caption lengths in shot and tactic captions.
    (b) Frequency distribution of the most common words in shot and tactic captions.}
    \label{fig:caption_analysis}
\end{figure}

\section{Shot2Tactic-Caption Dataset}
\subsection{Data Collection and Annotation}
\label{sec:data_collection}
In this study, we collected data from 10 men's doubles matches (a total of approximately 7h 37 min) in the BWF World Tour Super 1000 matches held in China and Malaysia in 2023 and 2024. 
All videos were collected from publicly available official YouTube broadcasts\cite{BWF_YouTube}.  
The dataset includes 5,494 shot video clips with an average duration of 0.7 seconds, and 544 tactic units with an average length of 5 seconds (150 frames), each annotated with a descriptive caption.

The data for each match is stored in individual JSON files and annotated with three types of information: (1) the filename of the match video along with the start and end timestamps and scores for each rally (i.e., a continuous exchange of shots between teams until the point ends); (2) the start and end times of each shot along with shot captions; and (3) tactic units each composed of multiple shots, annotated with their start and end times, tactic types, tactic states, and corresponding tactic captions.

Based on tactic definitions commonly used in badminton, this study treats a sequence of multiple shots as a single tactic segment and categorizes tactics into 9 types. ``Serve \& Attack'', ``Continuous Smashing'', ``Net Pressure \& Kill'', ``Push \& Trap'', ``Flick Serve Attack'', ``Push \& Smash'', ``Drop \& Net Domination'', ``Drive \& Intercept'', and ``Tempo Variation Control''.
For example, ``Continuous Smashing'' refers to an aggressive tactic aiming to break the opponent's defense by repeatedly applying pressure through consecutive smashes. A typical shot pattern is like:
(Smash, ?, Smash, ?, Smash).
The symbol ``?'' represents an arbitrary opponent shot. This abstraction reflects that the definition of the tactic does not rely on the specific type of the opponent’s response, as long as it allows the attacking player to continue the aggressive sequence.

In addition, each tactic unit consists of multiple shots, and each shot is annotated with a tactic state selected from: ``Start'', ``Continue'', ``Interrupt'', ``Resume'', and ``Finish''. These five states are designed to model the dynamic progression and possible disruption of tactical execution, inspired by the control flow in Transmission Control Protocol \cite{postel1981transmission}.

All annotations were conducted by seven annotators, each with at least 3 years of badminton experience and competitive amateur-level match experience.
For quality assurance, each rally was annotated independently by two annotators. Any disagreements were resolved through group discussion among all seven annotators, and the final label was determined by consensus.
Each caption was required to include the shot type, with the shot direction specified whenever it was clearly distinguishable (e.g., straight, crosscourt), and annotators were instructed to explicitly describe any changes in player formation.

For subsequent experiments, the dataset was split by match videos into 70\% for train, 10\% for validation, and 20\% for test, ensuring that there was no overlap of videos between the train and test sets.

\subsection{Dataset Statistics}

As shown in Figure~\ref{fig:caption_length_distribution}, the shot captions length is between 2 and 21 words, with an average of 5.8 words per caption. This concise style is motivated by the potential application in real-time tactical commentary scenarios. While the tactic captions length is between 43 and 98 words, with an average of 60.94 words per caption.

Figure~\ref{fig:most_common_words_cleaned} presents frequently used words in the shot and tactic captions. For shot caption, revealing a high occurrence of badminton-related verbs such as ``smash'', ``lift'', and ``push'', and nouns like ``net'', ``point'', and ``midcourt''. Many of these words reflect tactical concepts such as shot types and shuttle directions. Words like ``side'' and ``formation'', which describe positional formations in doubles play, also appear frequently.
For tactic caption, the most frequent words include ``sequence'', ``executing'', ``tactic'', ``responds'', ``successfully'', and ``finish''. These terms reflect a shift in focus from individual shot descriptions to higher-level tactical reasoning. Unlike shot captions, tactic captions tend to describe how tactics unfold, whether they are interrupted or resumed, and whether the tactic was successfully executed. Note that the dataset does not include player identification, and all players are uniformly referred to as ``[PLAYER]'' in the shot and tactic captions.

\begin{table}[t]
\centering
\caption{Tactic Categories and State Distributions.}
\label{tab:tactic_stats}
\small
% \begin{tabular}{l c @{\hspace{4em}} l c}
\begin{tabular}{l c | @{\hspace{2em}} l c}
\toprule
\textbf{Tactic Type} & \textbf{Count} & \textbf{Tactic State} & \textbf{Count} \\
\midrule
Serve \& Attack & 396 & Start     & 544 \\
Continuous Smashing & 52  & Continue  & 1512 \\
Net Pressure \& Kill & 20 & Interrupt & 440 \\
Push \& Trap & 13         & Resume    & 120 \\
Flick Serve Attack & 5    & Finish    & 544 \\
Push \& Smash & 11        &           &          \\
Drop \& Net Domination & 24 &        &          \\
Drive \& Intercept & 12   &           &          \\
Tempo Variation Control & 11 &       &          \\
\midrule
\textbf{Total} & \textbf{544} & & \\
\bottomrule
\end{tabular}
\end{table}

Table~\ref{tab:tactic_stats} summarizes the distribution of tactic categories and tactic states in the Shot2Tactic-Caption dataset. The left side lists the number of annotated samples for each tactic category, while the right side shows the overall frequency of each tactic state across all labeled units.
% In total, the dataset covers 9 tactic categories, including ``Serve \& Attack (396), ``Continuous Smashing'' (52), ``Net Pressure \& Kill'' (20), ``Push \& Trap'' (13), ``Flick Serve Attack'' (5), ``Push \& Smash'' (11), ``Drop \& Net Domination'' (24), ``Drive \& Intercept'' (12), and ``Tempo Variation Control'' (11), where the numbers in parentheses indicate the number of annotated samples for each tactic.

\begin{figure*}[t]
    \centering
    \includegraphics[width=0.8\textwidth]{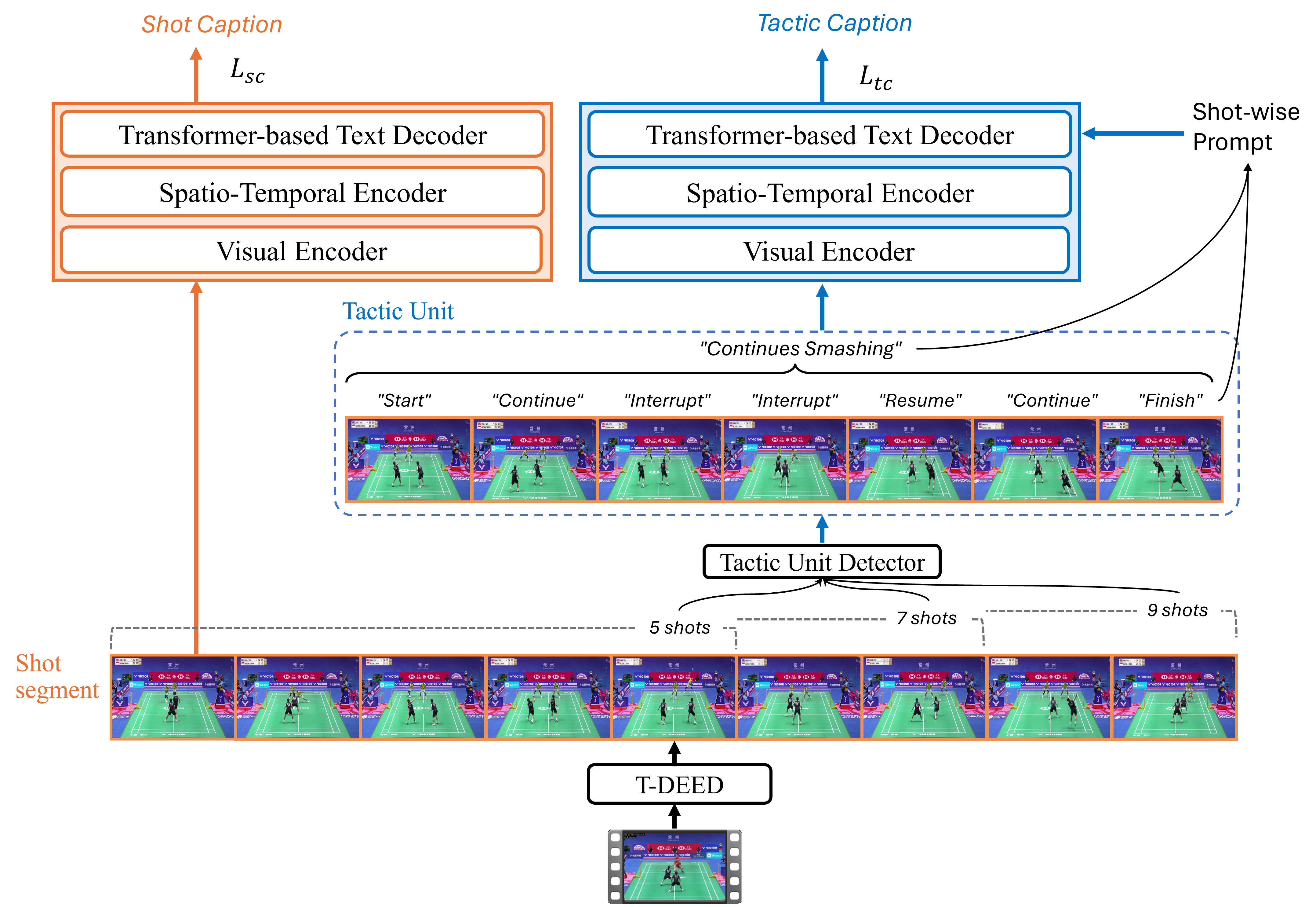}
    \caption{Overview of the Shot2Tactic-Caption framework. T-DEED\cite{xarles2024t} segments the input video into shots, which are then grouped into tactic units with predicted tactic type and state by the Tactic Unit Detector. Two parallel encoder-decoder branches generate shot captions (orange) and tactic captions (blue). The tactic branch leverages shot-wise prompts via cross-attention to guide tactic caption generation.}
    % \caption{Overview of the Shot2Tactic-Caption framework. Starting from the bottom, T-DEED\cite{xarles2024t} segments the input video into individual shots. These are grouped into tactic units by the Tactic Unit Detector, which also predicts tactic type and state. Based on this structure, two parallel encoder-decoder branches are employed: the shot captioning branch (orange) generates shot captions, while the tactic captioning branch (blue) generates higher-level tactic captions. Both branches share a similar architecture, consisting of a visual encoder (ResNet-50), a spatio-temporal Transformer encoder, and a Transformer-based text decoder. The tactic branch further incorporates a prompt-guided mechanism, where shot-wise prompts derived from the predicted tactic type and state are injected via cross-attention to guide tactical caption generation.}
    \label{fig:framework}
\end{figure*}

\section{Proposed Method: Shot2Tactic-Caption}
\subsection{Overall Architecture}

The whole architecture of Shot2Tactic-Caption is depicted in Figure~\ref{fig:framework}.
Our proposed method first applies T-DEED\cite{xarles2024t} to segment the input video into individual shot segments. Next, a sliding window approach is employed to construct candidate tactic segments, each consisting of 5, 7, or 9 consecutive shots. For each candidate, a Tactic Unit Detector is proposed to determine whether it constitutes a valid tactic sequence, to classify the tactic type, and identify the tactic states. A dual-branch architecture is employed to generate both shot and tactic captions.Details of the architecture are described in the following sections.

\subsection{Tactic Unit Detector}\label{sec:tactic_detection}

To support tactic captioning, we introduce a Tactic Unit Detector that operates in two stages. Given a candidate tactic segment composed of multiple shots (5, 7, 9 shots), the detector first predicts whether the segment corresponds to an ongoing tactic using a binary classifier.
If a segment is identified as a valid tactic, a second-stage classifier is applied to determine its specific tactic type and the current state within the progression. 
The detector is implemented as a lightweight video classification pipeline consisting of a ResNet3D-18 backbone\cite{tran2018closer}, a 2-layer Transformer encoder, and an attention pooling mechanism. We first apply a pretrained ResNet3D-18 to extract spatiotemporal features from each shot, where each shot consists of 16 uniformly sampled frames.
These per-shot feature sequences are then processed by a Transformer encoder with 512-dimensional embeddings and 8 attention heads. These shot-level embeddings are then averaged to produce a tactic-level representation.

In the first stage, this tactic-level representation is passed through a binary classification head to determine whether the segment corresponds to a valid tactic unit. If the segment is classified as valid, the same representation is forwarded to a second-stage classifier consisting of two parallel MLP heads: one predicts the tactic type (9-way classification), and the other predicts the current tactic state (5-way classification).

To encourage robust discrimination between visually similar segments and stabilize learning in early stages, we design a composite loss function $\mathcal{L}_{\text{detection}}$ consisting of a focal loss\cite{lin2017focal} and a negative margin penalty\cite{liu2020negative}:
\begin{equation}
\mathcal{L}_{\text{detection}} = \mathcal{L}_{\text{focal}} + \lambda \cdot \mathcal{L}_{\text{margin}}
\end{equation}
\noindent where $\lambda = 2$ balances the contribution of the margin penalty.
We compute the focal loss as:
\begin{equation}
% \text{FL}(p_t) = -\alpha_t (1 - p_t)^\gamma \log(p_t)
\mathcal{L}_{\text{focal}}  = -\alpha_t (1 - p_t)^\gamma \log(p_t)
\end{equation}
\noindent
where $p_t$ denotes the probability aligned with the ground truth, $\alpha_t$ balances class distribution, and $\gamma$ focuses learning on hard examples.

\noindent
To further reduce overconfident predictions on negative samples, we use a soft penalty term that penalizes predictions exceeding a gradually increasing threshold margin $m$:

\begin{equation}
\mathcal{L}_{\text{margin}} = \frac{1}{|\mathcal{N}|} \sum_{i \in \mathcal{N}} \max(0, \sigma(z_i) - m)
\end{equation}

\noindent
where $\mathcal{N}$ denotes the set of negative examples ($y_i = 0$), and $m$ is defined using a linear warm-up schedule: the margin $m$ is linearly increased from 0.1 to 0.5 over the first 5 epochs and then fixed at 0.5.
Here, $\sigma(\cdot)$ is the sigmoid function, and $z_i$ is the predicted logit for the $i$-th negative example.

For the second-stage classification model, which predicts the tactic type and states, we use a focal loss for the tactic type to address class imbalance, and a standard cross-entropy loss\cite{zhang2018generalized} for tactic states:

\begin{equation}
\mathcal{L}_{\text{type}} = -\sum_{k} \alpha_k (1 - \hat{y}_{\text{type},k})^\gamma y_{\text{type},k} \log \hat{y}_{\text{type},k}
\label{eq:focal_loss}
\end{equation}

\begin{equation}
\mathcal{L}_{\text{state}} = -\sum_{s} y_{\text{state},s} \log \hat{y}_{\text{state},s}.
\label{eq:ce_loss}
\end{equation}
\noindent
% where $y_{\text{type}}$ and $y_{\text{state}}$ are one-hot ground truth labels, and $\hat{y}$ denotes predicted probabilities. The class-wise weight $\alpha_k$ helps balance rare classes, and $\gamma$ is set as 2 to emphasize hard-to-classify examples.

\noindent
Here, $y_{\text{type}}$ and $y_{\text{state}}$ are one-hot ground truth labels, and $\hat{y}_{\text{type}}$ and $\hat{y}_{\text{state}}$ denote predicted probabilities.
Equation~\ref{eq:focal_loss} corresponds to the focal loss, where the class-wise weight $\alpha_k$ balances rare classes, and the focusing parameter $\gamma$ (set to 2) emphasizes hard-to-classify examples.
Equation~\ref{eq:ce_loss} is the standard cross-entropy loss used for tactic state prediction.

The total loss for tactic type and state classification is:
\begin{equation}
\mathcal{L}_{\text{classification}} = \mathcal{L}_{\text{type}} + \beta \cdot \mathcal{L}_{\text{state}}
\end{equation}
\noindent
where $\beta$ is set to 0.5 to balance the relative importance of the two objectives.

\subsection{Shot and Tactic Captioning Model Architecture}
 
Our model adopts an encoder-decoder architecture with two parallel branches: one for shot captioning (orange) and the other for tactic captioning (blue). 
Both the shot and tactic captioning branches use independent encoder-decoder pipelines. Each consists of a ResNet50-based\cite{he2016deep} visual encoder, a spatio-temporal encoder, and a Transformer-based text decoder.
Although both branches share the same structure, they are designed to capture different time scales and semantic objectives. The shot branch focuses on individual shots, while the tactic branch handles tactic unit composed of multiple shots. 

This encoder is designed to capture fine-grained spatio-temporal patterns. For each input frame, the ResNet-50 backbone extracts a high-resolution spatial feature map (typically of shape \( h \times w = 14 \times 14 \)), resulting in \( N = h \times w = 196 \) spatial locations per frame. These are projected to a 768-dimensional embedding and flattened into a sequence of spatio-temporal tokens across all \( T \) frames.
Instead of applying early spatial or temporal pooling, the encoder processes this flattened sequence directly with a Transformer. The model adds learnable positional encodings to represent both the temporal order across frames and the spatial position within each frame, allowing it to preserve detailed spatial relationships over time. This level of granularity enables the model to focus on subtle movements, such as wrist flicks, footwork, or shuttle changes, which are essential in badminton.

For tactic captioning, we divide the input tactic unit into multiple shots (5, 7, 9 shots), each consisting of 16 frames. 
Each shot is independently encoded into a sequence of tokens using a spatio-temporal encoder, and these sequences are then concatenated along the temporal axis to form a unified representation of the entire tactic sequence, which serves as input to the tactic text decoder.

For tactic captioning, we further incorporate a prompt-guided mechanism into the decoder. The decoder receives shot-wise prompts derived from predicted tactic type and state, which are injected via cross-attention to enhance context-aware caption generation. The full architecture is illustrated in Figure~\ref{fig:framework}.

\subsection{Shot Captioning}

After applying T-DEED \cite{xarles2024t} to segment the input video into individual shot segments, we proceed to generate captions for each shot. For each detected shot segment, we uniformly sample 16 consecutive frames. Each frame is passed through a ResNet-50 backbone to extract spatial features, which are then projected into a 768-dimensional embedding space. The resulting features are reshaped into spatio-temporal tokens and enriched with learnable spatial and temporal positional embeddings. These tokens are then processed by a Transformer-based decoder with 4 layers, 8 attention heads, which generates text descriptions using auto-regressive decoding.

The generated shot caption is trained with standard teacher forcing and by minimizing the cross-entropy loss between the predicted tokens and the ground truth shot captions. The shot captioning loss is computed as follows:
\begin{equation}
L_{\mathrm{sc}} = -\sum_{t=1}^{T} \log p(y_t \mid y_{<t}, \mathbf{H}^{\text{shot}}),
\end{equation}
where \(L_{\mathrm{sc}}\) denotes the shot captioning loss, \(y_t\) is the ground truth token at position \(t\), and \(\mathbf{H}^{\text{shot}}\)  is the encoded shot-level feature representation.

\subsection{Tactic Captioning}
For tactic captioning, we use the detected valid tactic segments described in Section~\ref{sec:tactic_detection} as input, each consisting of several consecutive shots.
Although the overall backbone architecture remains the same as in shot captioning, we adopt a different encoding strategy in the tactic captioning branch. Rather than encoding all frames of the entire tactic unit sequence as a whole, we first divide the input into individual shot segments and encode each shot independently using the same spatio-temporal encoder. Each shot segment contains 16 consecutive frames with resolution $3 \times 224 \times 224$. After feature extraction by the spatio-temporal encoder, each frame is converted into a $14 \times 14$ grid of patches (196 tokens), each represented by a 768-dimensional vector. The tokens from all 16 frames are then flattened into a sequence of 3136 tokens per shot to form the tactic-level representation. This design allows the model to preserve the internal motion dynamics of each shot while enabling the decoder to capture transitions and tactical dependencies across shots.

To guide the decoder with tactical context, we introduce a \textit{prompt-guided mechanism} based on the detected tactic type and state (Section~\ref{sec:tactic_detection}). These prompt labels are embedded with a 768-dimensional embedding layer and positional encoding, then injected via a single cross-attention layer before the 4-layer Transformer decoder.

We explore two prompt designs with different levels of structural alignment:
To incorporate temporal tactical cues during caption generation, we introduce a shot-wise prompt design. Given a tactic segment divided into \(N\) equal-length multiple shots (\(N=5, 7, 9\)), each shot is annotated with a tactic state label inferred by Tactic Unit Detector. With a global tactic type shared across the entire tactic unit, we construct the prompt as a sequence of state-aware descriptions:

\begin{center}
Prompt \(i\): <TacticType> -- <State\(_i\)>
\end{center}

These prompt entries are concatenated in temporal order to reflect the internal structure of the tactic progression. To retain temporal alignment, we apply shot-wise positional encoding and inject the resulting embeddings into the decoder via cross-attention.

To evaluate the impact of prompt structure, we design a baseline variant referred to as the flat prompt for comparison.  
In this variant, the entire tactic structure is collapsed into a single sentence without preserving the one-to-one correspondence between video segments and tactic states.  
For example:
\begin{center}
Prompt: <TacticType> -- <State\(_1\)> -- <State\(_t\)>
\end{center}
Here, \(t\) represents the index of the last state in a tactic unit.  
This flat prompt omits the temporal alignment between individual shots and their tactical roles.  
In contrast, our proposed shot-wise prompt explicitly models each shot’s tactical role, enhancing both interpretability and temporal precision in the generated captions.

The generated tactic caption is trained using standard teacher forcing by minimizing the cross-entropy loss between the predicted tokens and the ground truth tactic-level annotations. The tactic segment captioning loss is computed as follows:
\begin{equation}
L_{\mathrm{tc}} = -\sum_{t=1}^{T} \log p(y_t \mid y_{<t}, \mathbf{H}^{\text{tactic}}),
\end{equation}
where \(L_{\mathrm{tc}}\) denotes the tactic captioning loss, \(y_t\) is the ground truth token at position \(t\), and \(\mathbf{H}^{\text{tactic}}\)  is the encoded tactic-level feature representation.

% flat prompt and shot-wise prompt. The flat prompt encodes all tactical states as a single sentence, without explicit alignment to shot segments. In contrast, the shot-wise prompt maintains the sequential structure by aligning each state with its corresponding video segment.
\begin{table}[t]
\centering
\small  % 字体变小
\setlength{\tabcolsep}{0.5pt}  % 减小列间距
\caption{Performance of the Tactic Unit Detector.}
\label{tab:detector_results}
\begin{tabular}{lcccc}
\toprule
Component & Accuracy (\%) & Macro-F1 (\%) & Precision (\%) & Recall (\%) \\
\midrule
Valid Tactic Unit (Binary) & 89.28 & 84.18 & 78.70 & 73.32 \\
Tactic Type (9 classes) & 81.27 & 74.68 & -- & -- \\
Tactic State (5 classes) & 80.63 & 79.34 & -- & -- \\
\bottomrule
\end{tabular}
\end{table}

\begin{table*}[t]
    \centering
    \caption{Evaluation results comparison for shot and tactic captioning. 
    \textit{Note:} All models take multiple frames as input, with a size of \(224 \times 224\) due to the constraints of pretrained weights.}
    \label{tab:metrics_both}
    \resizebox{\textwidth}{!}{
    \begin{tabular}{c c c|ccccccccc}
        \toprule  % 最上面的粗横线

        \multirow{2}{*}{Granularity} & \multirow{2}{*}{Encoder} & \multirow{2}{*}{Decoder} &
        \multicolumn{9}{c}{Evaluation Metrics} \\ \cmidrule(l){4-12}
        & & & BLEU-1 & BLEU-2 & BLEU-3 & BLEU-4 & CIDEr & METEOR & ROUGE-L & Precision & Recall \\ 
        \midrule

        % --- Shot-level rows ---
        \multirow{5}{*}{Shot} 
            & ResNet50-STE & Transformer & \textbf{65.42} & \textbf{61.22} & \textbf{56.54} & \textbf{48.99} & \textbf{26.22} & \textbf{70.78} & \textbf{57.67} & 70.58 & \textbf{66.60}  \\
            & TimeSformer\cite{bertasius2021space} & Transformer & 59.75 & 55.99 & 51.45 & 43.50 & 20.26 & 66.64 & 52.64 & 70.29 & 60.47 \\
            & \multicolumn{2}{c|}{SoccerNet-Caption~\cite{xarles2024t}} & 53.76 & 48.43 & 43.15 & 35.00 & 20.10 & 66.30 & 47.64 & 53.76 & 63.62 \\
             % & \multicolumn{2}{c|}{Video-LLaMA\cite{damonlpsg2023videollama}} & 51.17 & 48.94 & 45.92 & 39.41 & 18.20 & 64.16 & 53.45 & 80.51 & 55.40 \\
            & \multicolumn{2}{c|}{VideoLLaMA2\cite{damonlpsg2024videollama2}} & 51.17 & 48.94 & 45.92 & 39.41 & 18.20 & 64.16 & 53.45 & \textbf{80.51} & 55.40 \\
            % & \multicolumn{2}{c|}{mPLUG-Owl2\cite{ye2023mplugowl2}} & 27.84 & 14.06 & 6.06 & 2.92 & 0.56 & 26.74 & 36.78 & 40.48 & 29.45 \\
            & \multicolumn{2}{c|}{Qwen2.5-VL\cite{Qwen2.5-VL}} & 58.11 & 51.93  & 45.64  & 35.52  & 14.84 & 60.97 & 43.03 & 76.10 & 58.25  \\
        \midrule

        % --- Tactic-level rows ---
        \multirow{5}{*}{Tactic} 
            & ResNet50-STE & Transformer & \textbf{75.86} & \textbf{71.51} & \textbf{68.03} & \textbf{64.57} & 17.79 & \textbf{79.63} & \textbf{81.54} & \textbf{90.14} & 76.88\\
            & TimeSformer\cite{bertasius2021space} & Transformer & 75.14 & 70.48 & 66.80 & 63.09 & 13.42 & 78.39 & 79.43 & 88.61 & 76.07 \\
            % &
            & \multicolumn{2}{c|}{SoccerNet-Caption~\cite{xarles2024t}} & 74.38 & 68.57 & 64.19 & 59.88 & 14.43 & 77.94 & 76.87 & 84.97 & 75.84 \\
            % & TimeSformer\cite{bertasius2021space} & Transformer & 59.75 & 55.99 & 51.45 & 43.50 & 2.03 & 66.64 & 52.64 & 70.29 & 60.47 \\
             % & \multicolumn{2}{c|}{Video-LLaMA\cite{damonlpsg2023videollama}} & 73.74 & 69.03 & 65.13 & 61.30 & \textbf{15.75} & 76.46 & 75.13 & 85.28 & 74.45 \\
            & \multicolumn{2}{c|}{VideoLLaMA2\cite{damonlpsg2024videollama2}}  & 73.74 & 69.03 & 65.13 & 61.30 & \textbf{17.85} & 76.46 & 80.13 & 89.28 & 74.45 \\
            % & \multicolumn{2}{c|}{mPLUG-Owl2\cite{ye2023mplugowl2}} & 27.84 & 14.06 & 6.06 & 2.92 & 0.56 & 26.74 & 36.78 & 40.48 & 29.45 \\
            & \multicolumn{2}{c|}{Qwen2.5-VL\cite{Qwen2.5-VL}} & 73.47 & 67.02  & 62.24 & 57.35 & 11.27 & 78.54 & 78.15 & 73.47 & \textbf{81.06}   \\
        \bottomrule  % 最下面的粗横线
    \end{tabular}}
\end{table*}

% \begin{table*}[t]
% \centering
% \caption{Captioning performance using different prompt types. Each setting reports results for both shot-level and tactic-level captioning.}
% \label{tab:prompt_type_results}
% \small
% \setlength{\tabcolsep}{6pt}
% \renewcommand{\arraystretch}{1.1}
% \begin{tabular}{l l | c c c c c c}
% \toprule
% Prompt Type & Granularity & BLEU-4 & CIDEr & METEOR & ROUGE-L & Precision & Recall \\
% \midrule
% \multirow{2}{*}{Flat} 
%   & Shot   & 48.43 & 25.86 & 70.38 & 57.48 & 72.13 & 64.84 \\
%   & Tactic & 63.23 & 16.14 & 75.12 & 77.88 & 80.70 & 76.72 \\
% \midrule
% \multirow{2}{*}{Shot-wise} 
%   & Shot   & 48.99 & 26.22 & 70.78 & 57.67 & 70.58 & 66.60 \\
%   & Tactic & 64.57 & \textbf{17.79} & 79.63 & 81.54 & \textbf{90.14} & 76.88 \\
% \bottomrule
% \end{tabular}
% \end{table*}

\begin{table*}[t]
\centering
\caption{Captioning performance using different prompt types. Each setting reports results for both shot and tactic captioning.}
\label{tab:prompt_type_results}
\small
\setlength{\tabcolsep}{6pt}
\renewcommand{\arraystretch}{1.1}
\begin{tabular}{l l | c c c c c c}
\toprule
Prompt Type & Granularity & BLEU-4 & CIDEr & METEOR & ROUGE-L & Precision & Recall \\
\midrule
\multirow{2}{*}{w/o prompt} 
  & Shot   & 47.92 & 25.06 & 69.91 & 56.72 & 71.43 & 64.33 \\
  & Tactic & 59.45 & 14.53 & 72.85 & 72.25 & 74.37 & 74.13 \\
\midrule
\multirow{2}{*}{Flat} 
  & Shot   & 48.43 & 25.86 & 70.38 & 57.48 & 72.13 & 64.84 \\
  & Tactic & 63.23 & 16.14 & 75.12 & 77.88 & 80.70 & 76.72 \\
\midrule
\multirow{2}{*}{Shot-wise} 
  & Shot   & 48.99 & 26.22 & 70.78 & 57.67 & 70.58 & 66.60 \\
  & Tactic & 64.57 & \textbf{17.79} & 79.63 & 81.54 & \textbf{90.14} & 76.88 \\
\bottomrule
\end{tabular}
\end{table*}

\subsection{Captioning Loss Function}

To jointly train the shot and tactic captioning branches, we adopt a weighted multi-task loss and \( \mathcal{L}_{\text{toal}} \) is computed as below:

\begin{equation}
\mathcal{L}_{\text{total}} = \lambda_{\text{sc}} \cdot \mathcal{L}_{\text{sc}} + \lambda_{\text{tc}} \cdot \mathcal{L}_{\text{tc}},
\end{equation}

\noindent where \( \mathcal{L}_{\text{sc}} \) and \( \mathcal{L}_{\text{tc}} \) denote the cross-entropy losses for clip-level and tactic-level caption generation, respectively.

We set the loss weights to \( \lambda_{\text{sc}} = 0.3 \) and \( \lambda_{\text{tc}} = 6 \) to emphasize the learning of tactic captioning while retaining the benefits of shot caption supervision. This reflects the fact that tactic captioning requiring stronger guidance during training.
Although the two branches are architecturally independent, joint training leverages the complementary nature of the tasks and the shared embedding space to enhance overall modeling capacity and generalization performance. The Tactic Unit Detector is trained separately and not included in the captioning loss.

\section{Experiments and Results}
\subsection{Experimental Settings}

Each video frame was resized to \(224 \times 224\) before being input to the model. For shot captioning, a sequence of 16 frames, uniformly sampled from the original video was used. For tactic-level captioning, tactic segments consisted of 5 shots (80 frames), 7 shots (112 frames), or 9 shots (144 frames). The segment length was determined based on the definition of minimal tactical units such as ``Continuous Smashing'', which requires at least three successive smashes interleaved with opponent returns. Shots that do not match the defined shot types are considered interruptions, and up to two interruptions are allowed within a segment in this study.  For example, a sequence like: (Smash, ?, Lift, ?, Smash, ?, Smash) would still be considered a valid ``Continuous Smashing'' tactic unit because the ``Lift'' is treated as an interruption but does not exceed the allowed threshold.
Thus, the total tactic sequence length can vary from 5 to 9 shots depending on the occurrence of such interruptions.

In this study, we trained our models for 30 epochs using the AdamW optimizer with a learning rate of  \(1 \times 10^{-5}\), a weight decay of 0.01, and a batch size of 4. To improve training stability, we employed a linear learning rate scheduler with a warm-up phase comprising the first 10\% of the total training steps.
The models were initialized with pretrained weights. Specifically, ResNet50-STE was fine-tuned with weights pretrained on ImageNet \cite{5206848} . TimeSformer \cite{bertasius2021space} was initialized with weights pretrained on Kinetics-400 \cite{kay2017kinetics}. Video-LLaMA2 \cite{damonlpsg2024videollama2} and Qwen2.5-VL \cite{Qwen2.5-VL} were fine-tuned with 7B pretrained language model weights.

\subsection{Experimental Results}

\subsubsection{Tactic Unit Detector}

For model training, we used a total of 1,410 labeled segments, comprising 544 valid tactic units and 866 non-tactic segments. 
The data was split according to the strategy described in Section~\ref{sec:data_collection}.
The performance of the Tactic Unit Detector is summarized in Table~\ref{tab:detector_results}. 
The binary classification task for detecting the presence of tactics achieves relative high accuracy, demonstrating the model’s capability in distinguishing tactic unit from non-tactic unit.
In comparison, the multi-class classification tasks for predicting tactic types (9 classes) show relatively lower performance. The tactic type classifier achieves an accuracy of 81.27\% and a Macro-F1 score of 74.68\%. The relatively lower Macro-F1 score for tactic type classification likely reflects the imbalanced distribution of tactic categories in our dataset. Specifically, certain tactics (e.g., Flick Serve \& Attack) occur much less frequently than others, making it more challenging for the model to learn reliable patterns for these rare tactic types.

\begin{figure*}[t]
\centering
\includegraphics[width=0.9\textwidth]{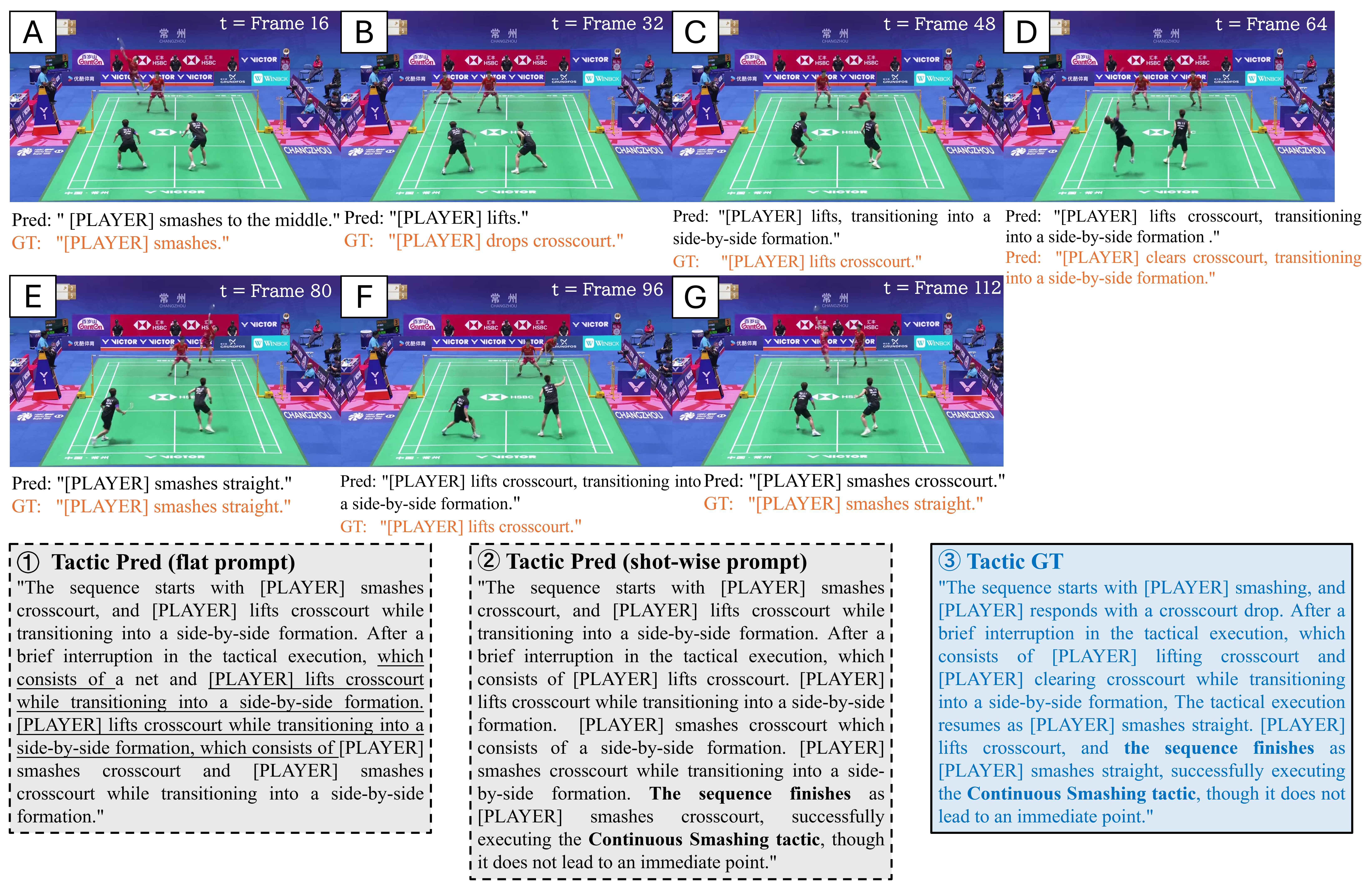}
\caption{Qualitative examples of shot captions (A--G) and tactic captions (\ding{172}--\ding{174}). Different prompt types are used in tactic captioning to illustrate their effect on description quality.}
\label{fig:qualitative}
\end{figure*}

\subsubsection{Ablation study on Encoders and Decoders}
Table~\ref{tab:metrics_both} shows the performance comparison of caption generation models at the shot level and tatic-level. We evaluated different combinations of encoders and decoder, including ResNet50-based spatio-temporal encoder (ResNet50-STE), TimeSformer encoder, SoccerNet-Caption, VideoLLaMA2 and Qwen2.5-VL. Although SoccerNet-Caption, VideoLLaMA2, and Qwen2.5-VL are not strictly modular encoder-decoder models, we employ their full captioning pipelines as alternatives to our encoder-decoder setup for comparison purposes.

Among these, use ResNet50-STE as Encoder achieved the highest scores across nearly all metrics. For shot captioning, it reached a BLEU-4\cite{10.3115/1073083.1073135} of 48.99 and a METEOR\cite{banerjee2005meteor} of 70.78, outperforming the second-best TimeSformer by 5.5 and 4.1 points respectively. Similarly, in tactic-level captioning, ResNet50-STE achieved a BLEU-4 of 64.57 and a METEOR of 79.63, exceeding the one takes TimeSformer as Encoder by 1.5 and 1.2 points. While VideoLLaMA2 achieved the highest CIDEr score\cite{Vedantam_2015_CVPR} (17.85) in tactic-level results which suggesting that it generates more diverse or detailed descriptions. Qwen2.5-VL achieves the highest Recall (81.06\%) in tactic-level captioning. This suggests that it captures a broader range of relevant content.
These results highlight that badminton videos, characterized by relatively stable backgrounds and subtle motion patterns, benefit from preserving fine-grained spatial details in each frame, which can be effectively captured by a ResNet-based encoder combined with temporal modeling.

\subsubsection{Ablation Study on Prompt Structure}
We compared three different prompt settings: w/o prompt, flat prompt, and shot-wise prompt. The w/o prompt setting removes any explicit prompt information from the input, allowing the model to generate tactic captions purely from visual features. This serves as a baseline to assess how much structured prompts can guide the captioning process.
It should be noted that prompts are only used for tactic caption generation. However, since the model is trained jointly for both shot and tactic captioning, modifications to the prompt can indirectly influence the results of both tasks. 
As shown in Table~\ref{tab:prompt_type_results}, the shot-wise prompt outperforms the flat prompt and w/o prompt on both shot and tactic captioning.  For shot captioning, shot-wise prompting slightly improves CIDEr (from 25.86 to 26.22) and Recall (from 64.84 to 66.60), showing better informativeness and coverage.
More notable improvements are observed in tactic captioning: CIDEr increases from 16.14 to 17.79, METEOR from 75.12 to 79.63, and Precision from 80.70 to 90.14. This indicates that maintaining the order and structure of tactical segments helps the model better understand the temporal flow of a tactic, resulting in more accurate and focused descriptions.

These results suggest that preserving the sequential structure of tactical cues, rather than flattening them into a single sentence, is beneficial for both shot and tactic caption generation.

\subsubsection{Qualitative Results}

Figure~\ref{fig:qualitative} presents qualitative examples of predicted captions for shot and tactic unit. These examples illustrate both the strengths and limitations of our model. As shown in Figure~\ref{fig:qualitative}, the proposed model can provide highly accurate captions for individual shots and produce coherent tactic-level descriptions in most cases.

For shot captioning, the model successfully captures the core action in many instances. For example, when the ground truth is ``[PLAYER] smashes'', the model correctly generates "[PLAYER] smashes to the middle", which preserves the action verb and enriches it with additional spatial detail (Figure~\ref{fig:qualitative}A). 
However, there are also cases where the model fails to identify shot types or directions. For example, it may generate “[PLAYER] lifts” instead of the ground truth “[PLAYER] drops crosscourt,” likely due to the visual similarity between the two shot types (Figure~\ref{fig:qualitative}B). These results suggest that while the model has learned general shot types and some spatial context, it still struggles with distinguishing subtle shot variations, particularly in fast-paced or ambiguous situations.
%
% ``Continuous Smashing''
For tactic captioning, we further investigate the impact of prompt design, we compare the predicted tactic captions using flat prompt and shot-wise prompt.

As shown in \ding{172} at the bottom of Figure~\ref{fig:qualitative}, the tactic caption predicted using a flat prompt exhibits repetition (underlined) and fails to include the tactic type (``Continuous Smashing'') and conclude the tactic unit (``the sequence finishes as...'' as shown in Figure~\ref{fig:qualitative} \ding{174}). Although the flat prompt contains tactic type and state information, its unstructured format prevents the model from effectively utilizing it to produce coherent descriptions or identify the tactic type.
In contrast, as shown in \ding{173} at the bottom of Figure~\ref{fig:qualitative}, the tactic caption predicted using a shot-wise prompt performs better in multiple aspects. First, the overall temporal structure is better captured—both the beginning and ending of the tactical sequence are clearly defined (e.g., “The sequence finishes as...”), and transitions between individual shots are logically coherent. Second, the model correctly identifies the tactic type (e.g., “Continuous Smashing tactic”), which is enabled by the structured, temporally aligned guidance provided in the shot-wise prompt.

Overall, shot-wise prompt provides clearer guidance to the decoder, resulting in more fluent and structured outputs. This finding supports our design choice of using shot-wise prompt encoding for tactic captioning.

\vspace{-5pt}
\section{Conclusion}

This study proposes a novel framework, Shot2Tactic-Caption, capable of generating both shot and tactic captions for badminton.  
To support this task, we also introduce the Shot2Tactic-Caption Dataset, the first badminton doubles captioning dataset.
In addition, we incorporate predefined tactic states to model the dynamic progression and potential disruptions in tactical execution.  
Furthermore, our method leverages a shot-wise prompt-guided mechanism to capture the temporal flow of tactics, to capture the temporal flow of tactics, allowing the model to more accurately reflect the tactical progression over time.
Future work includes conducting human evaluation to ensure that the quality of generated captions is assessed more reasonably and aligns with expert tactical understanding, extending the proposed framework to additional racket sports, such as table tennis and tennis, and examining its generalization across different game dynamics.
We also plan to explore downstream applications of the generated captions, including tactical information retrieval, visual question answering (VQA), and advanced performance analysis to support data-driven coaching and strategy development.

\begin{acks}
This work was financially supported by JST, ACT-X Grant Number JPMJAX24CH, and JSPS KAKENHI Grant Number JP24K23889.
The authors would also like to thank Prof. Shin'ichi Satoh for his valuable guidance and support.
We also express our sincere gratitude to students from Wuhu Institute of Technology, China, for their contributions to the data annotation process.
\end{acks}
\bibliographystyle{ACM-Reference-Format}
\balance
\bibliography{sample-base}

% \appendix









\end{document}